# Is AI Catching Up to Human Expression? Exploring Emotion, Personality, Authorship, and Linguistic Style in English and Arabic with Six Large Language Models

Nasser A Alsadhan

College of Computer and Information Sciences, King Saud University, Riyadh, Saudi Arabia
nsadhan@ksu.edu.sa

## Abstract

The advancing fluency of large language models (LLMs) raises important questions about their ability to emulate complex human traits, including emotional expression and personality, across diverse linguistic and cultural contexts. This study investigates whether state-of-the-art LLMs can convincingly mimic emotional nuance in English and personality markers in Arabic, a critical under-resourced language with unique linguistic and cultural characteristics. We conduct two tasks across six models—Jais, Mistral, LLaMA, GPT-4o, Gemini, and DeepSeek. First, we evaluate whether machine classifiers can reliably distinguish between human-authored and AI-generated texts. Second, we assess the extent to which LLM-generated texts exhibit emotional or personality traits comparable to those of humans.

Our results demonstrate that AI-generated texts are distinguishable from human-authored ones (F1 > 0.95), though classification performance deteriorates on paraphrased samples, indicating a reliance on superficial stylistic cues. Emotion and personality classification experiments reveal significant generalization gaps: classifiers trained on human data perform poorly on AI-generated texts and vice versa, suggesting LLMs encode affective signals differently from humans. Importantly, augmenting training with AI-generated data enhances performance in the Arabic personality classification task, highlighting the potential of synthetic data to address challenges in under-resourced languages. Model-specific analyses show that GPT-4o and Gemini exhibit superior affective coherence, while LLaMA performs weaker.

Linguistic and psycholinguistic analyses reveal measurable divergences in tone, authenticity, and textual complexity between human and AI texts. These findings have significant implications for affective computing, authorship attribution, and responsible AI deployment, particularly within under-resourced language contexts where generative AI detection and alignment pose unique challenges.

# 1. Introduction

The adoption of LLMs in domains such as education, marketing, legal drafting, and mental health has expanded rapidly (Hadi et al, 2023; Pearson, 2024; Krook et al., 2024; Taşkıran et al., 2025; Lang et al., 2024). With generative tools like GPT-4 and Gemini producing highly fluent text, concerns are rising around their ability to emulate human-like emotional tone and individual stylistic voice (Olteanu et al., 2024). But does fluency equate to authenticity? Can these models replicate not just what we say, but how we feel and who we appear to be?

Understanding whether LLMs can convincingly mimic human traits has far-reaching implications. In fields like psychotherapy, education, and authorship evaluation, such models may inadvertently shape or misrepresent affective states and individual identity. For instance, therapeutic chatbots may appear empathic without a genuine understanding of emotional nuance, potentially affecting user trust and mental health outcomes. Similarly, in educational settings, undetected AI-generated student work could disrupt assessment fairness and blur boundaries of authorship and originality. Beyond practical deployments, the expressive capacity of LLMs raises critical ethical concerns regarding deception, emotional manipulation, and authenticity in digital communication.

These issues underscore the urgency of empirical evaluation. If LLMs can imitate emotional or personal style convincingly, it challenges the ability of both humans and automated systems to differentiate synthetic from human-authored content. Conversely, if such models fail to achieve expressive realism, their utility in high-stakes applications must be reconsidered. Clarifying these capabilities is not only methodologically relevant but essential for informing AI deployment policies, authorship detection tools, and ethical design frameworks.

This study targets two expressive traits: emotional expression in English and personality traits in Arabic. These linguistic features vary structurally, emotions are often signaled through affective vocabulary and polarity (Alsadhan and Skillicorn, 2020), while personality traits tend to manifest in more subtle markers such as syntax, function words, and self-referential language (Paech et al., 2025; Abbasian et al., 2024; Roy et al., 2024). The ability of LLMs to convincingly emulate these signals raises key concerns: if these models can generate expressive content that mimics human language closely enough, it may complicate authorship attribution and blur the line between synthetic and human-authored text in evaluative settings like essays, therapy transcripts, or narrative writing (Ratican & Huston, 2024).

To investigate this, we conduct a comparative evaluation of six state-of-the-art LLMs: GPT-4 (Gallifant, 2024), Gemini (Team G et al., 2023), Mistral (Jiang et al., 2023), Jais (Sengupta et al., 2023), LLaMA 3 (Grattafiori et al., 2024), and DeepSeek (Liu et al., 2024). The study is organized around two core tasks: (1) distinguishing between human- and LLM-generated texts, and (2) evaluating how well these models replicate stylistic features through both machine learning classification and psycholinguistic profiling. We also introduce paraphrased versions of LLM outputs to test whether surface-level transformations can mask algorithmic signatures and increase perceived realism.

Arabic introduces an additional challenge to this evaluation. As a morphologically rich and dialectally diverse language, it poses unique representational difficulties for current LLMs, many of which are trained on limited Arabic corpora (Alshammari et al. 2024; Alhayan et al., 2024; Alsadhan, 2025). Testing personality expression in Arabic therefore serves as a crucial cross-linguistic probe, one that examines whether stylistic mimicry generalizes across cultural and structural boundaries.

While existing work has begun exploring LLM expressiveness, many studies limit their focus to a single language, model, or analytic method, typically English outputs from GPT or Gemini (Lang et al., 2024; Abbasian et al., 2024). In contrast, this study offers a broader empirical comparison. By evaluating six contemporary LLMs, some multilingual, others region-specific, across both English and Arabic expressive tasks, which contributes to a richer understanding of model behavior. Emotional expression is analyzed in English, while personality traits are analyzed in Arabic. This dual approach allows us to capture linguistic and cultural variability, often ignored in earlier benchmarks. Further, the inclusion of paraphrased outputs across all models allows for testing generalization and deception under more realistic conditions. Taken together, the study offers a structured evaluation of existing systems' expressive realism, a concern gaining urgency in AI ethics, education, and authorship attribution. Its contribution is therefore empirical and comparative, advancing the field's understanding of whether and how current LLMs convincingly mimic human emotion and personality across linguistic and cultural boundaries.

The rest of the paper is organized as follows. Section 2 reviews related work on generative AI's ability to emulate human emotion and personality, including authorship detection and linguistic feature analysis. Section 3 describes our proposed methodology, including datasets, experimental pipelines, and evaluation metrics. Section 4 presents and discusses the results, covering authorship attribution, emotion and personality classification, LLM-level performance variation, and linguistic feature analysis. Finally, Section 5 provides the conclusion and outlines directions for future research.

## 2. Literature Review

Recent advancements in LLMs have prompted researchers to explore not only their functional capabilities, but also their ability to emulate human traits such as emotion, personality, and stylistic nuance. While foundational surveys, such as (Hadi et al., 2023), provide comprehensive overviews of LLM architectures and applications, there is a growing need to examine how these models simulate human-like language across affective and cultural dimensions. This literature review integrates findings from recent studies focused on emotional tone, personality, stylometry, and psycholinguistic traits in generative AI, with particular attention to English and Arabic texts,

Linguistic Inquiry and Word Count (LIWC) based analysis (Pennebaker et al., 2001), and machine-classifier evaluation frameworks.

Paech (2023) introduced EQ-Bench, a benchmark designed to assess emotional intelligence in LLMs using multiple-choice social reasoning tasks. Though it offers a robust ranking system with high correlation to MMLU (r = 0.97), it is limited by its exclusive use of English and lacks breakdowns by emotion category. Complementing this, Roy et al. (2024) proposed ECGText, combining GPT-2-based generation with a RoBERTa emotion classifier. Evaluated on diverse domains (e.g., reviews, speeches), ECGText outperformed CTGAN on cosine similarity and POS-distribution metrics, demonstrating enhanced affective alignment through hybrid generation-classification strategies.

Stylometric and psycholinguistic analysis continue to play a vital role in evaluating human-likeness in text. The LIWC dictionary developed by Pennebaker et al. (2001) remains a foundational tool for mapping lexical features into psychological categories (e.g., affect, cognition, function words). Mikros (2025) examined GPT-4o's ability to imitate the literary styles of authors such as Hemingway and Shelley using stylometric analysis. The study found that while GPT-4o could approximate certain surface-level stylistic features, such as sentence and word length, its outputs remained distinguishable from human-authored texts, while Dubey (2024) found that LLM-derived stylistic embeddings outperformed traditional features in authorship classification, achieving 82.47% accuracy on unseen authors. In addition, cluster analysis confirmed that the stylistic embeddings captured authorial style over topic content, supporting their value for profiling linguistic signatures and detecting synthetic text. Opara (2024) proposed StyloAI, which uses 31 handcrafted features and Random Forest classifiers to distinguish AI from human text, achieving F1 scores of 0.97 across domains.

Beyond stylistic accuracy, understanding personality in LLMs has led to the development of purpose-built benchmarks. Lee et al. (2024) introduced TRAIT, an 8,000-item benchmark combining Big Five and Dark Triad frameworks enriched with ATOMIC10x commonsense scenarios. TRAIT outperforms traditional self-assessment surveys in content validity, reliability, and refusal rate. Results show that GPT-4 exhibits higher agreeableness and conscientiousness post-alignment, while prompting alone fails to induce certain traits (e.g., psychopathy). Likewise, Li et al. (2024) introduced the EERPD framework, which fuses emotional and emotion-regulation features with language models to predict personality, offering an expanded view of affective personality modeling. Dubey (2024) and Huang et al. (2025) both highlight the growing robustness of authorship attribution frameworks in the LLM era.

In Arabic NLP, research has recently expanded beyond translation and classification to include emotion and personality modeling. Dandash & Asadpour (2025) curated AraPers, an Arabic dataset for personality classification using BERT and LSTM, reporting >74% accuracy. El Bahri et al. (2024) analyzed Big Five traits in social learning contexts. Essam and Abdo (2021) used LIWC to analyze Arabic tweets during the COVID-19 pandemic, revealing dominant categories such as sadness, religion, and conspiracy, reinforcing LIWC's applicability in Arabic psycholinguistic contexts.

Authors such as Sengupta et al. (2023) introduced Arabic-centric models like Jais and Jais-Chat, which performed strongly on QA and classification tasks across BLEU, F1, and EM metrics. Other work includes Alhayan and Himdi (2024), who built an ensemble classifier combining CNN and logistic regression, reaching 89.70% accuracy in detecting AI-generated Arabic reviews. Nfaoui and Elfaik (2024) benchmarked ChatGPT across fine-tuning, in-context learning, and prompting for Arabic emotion recognition, finding fine-tuning most effective. Recent work in detection includes Alshammari et al. (2024) using contextual embeddings and a diacritization filter for Arabic AI-generated text detection in essays, achieving 98.4% accuracy.

Evaluations of generative models for content creation further inform discussions of human-likeness. Lang et al. (2024) compared GPT-4 and Gemini in crafting educational case studies, using readability indices and various linguistic metrics. GPT-4 outperformed Gemini on subjective measures. In multi-author contexts, Zamir et al. (2024) demonstrated how stylometric methods can identify author shifts and composite styles, which has implications for detecting LLM-generated composite texts. Klinkert et al. (2024) examined how LLMs emulate human-like personalities in non-Player Characters (NPCs) using International Personality Item Pool (IPIP) and Big Five frameworks, finding that GPT-4 achieved more coherent and human-aligned traits than other models. Lee et al. (2024) used psychotherapy-based prompting to enhance empathy in GPT-3.5-Turbo responses. Ferrario et al. (2024) assessed LLM behavior in depression-related interactions, noting fluency and empathy but also prompt instability and contextual limitations.

Huang et al. (2024) systematically compared LLM authorship inference across genres, showing that stylistic consistency persists despite model size, though detection remains challenging in highly polished outputs. Similarly, Opara (2024) and Mikros (2025) reinforced that stylometric classifiers still offer high performance in AI detection, especially when linguistic features are paired with ensemble models.

Despite promising outcomes, the literature reveals several persistent limitations. Most studies remain monolingual, focused on English, and evaluate a narrow set of popular models (e.g., GPT-3.5, GPT-4). Emerging models like Gemini, LLaMA, and DeepSeek are often underrepresented, and cross-lingual and multicultural perspectives are sparse. Evaluation strategies vary, with many relying on subjective or psychometric tests while overlooking classifier-based or feature-engineered methods. Moreover, emotional and personality traits are often analyzed in isolation, neglecting their real-world entanglement.

Our study addresses these gaps by presenting a comprehensive comparative evaluation of emotion and personality expression in LLMs across English and Arabic. Without relying on fine-tuning or model-specific modification, we systematically analyze emotion generation in English and personality generation in Arabic, comparing six major LLMs under consistent experimental conditions. In addition to classifier-based detection, we investigate whether paraphrasing AI-generated text affects authorship attribution, shedding light on the robustness of detection frameworks. We further contribute a detailed psycholinguistic analysis using LIWC, exploring how LLM outputs align with or diverge from human-authored linguistic profiles. By combining neural classification and feature-based diagnostics, our study highlights both shallow and deep stylistic cues and introduces a multilingual benchmark dataset to support future research on generative models' stylistic and psychological realism across languages, traits, and cultural contexts.

# 3. Methodology

This study presents a comparative framework to investigate whether LLMs can emulate human linguistic expression across two key traits: emotion (in English) and personality (in Arabic). This dual-language approach is designed to leverage high-quality, fine-grained human benchmarks in English while addressing the need for Modern Standard Arabic (MSA) expressive datasets. Without fine-tuning or modifying the LLMs, we use them solely to generate synthetic data via constrained prompting for evaluation. We approach this as a text classification problem, combining fine-tuned neural classifiers, BERT for English (Devlin et al., 2019) and AraBERT for Arabic (Antoun et al., 2020 for Arabic), with feature-based linguistic analysis.

Our methodology consists of three experimental pipelines: (1) AI Authorship Detection, (2) Emotion and Personality Classification across human and synthetic training regimes, and (3) Linguistic Feature Analysis. Each pipeline is applied separately to English and Arabic to ensure results are interpreted within their respective linguistic and cultural contexts. Together, these pipelines provide a comprehensive evaluation of both the detectability and stylistic realism of LLM-generated text, integrating neural classification performance with detailed psycholinguistic feature profiling.

## 3.1 Datasets

We utilized four publicly available emotion datasets in English (Wallace et al., 2014; Tanyel, 2022; Cheela, 2023; Elgiriyewithana, 2024) and one personality dataset in Arabic (Chraibi et al., 2024). The English datasets encompass ten emotion categories (sadness, joy, anger, fear, surprise, sarcasm, irony, disgust, trust, and anticipation), while the Arabic dataset consists of Modern Standard Arabic (MSA) texts annotated according to the Big Five personality traits: Openness, Conscientiousness, Extraversion, Agreeableness, and Neuroticism (Roccas et al., 2002). At the time of data collection, publicly available Arabic emotion datasets were predominantly dialectal, whereas the personality dataset used in this study was newly released and annotated in MSA, making it particularly suitable for controlled linguistic analysis.

The decision to evaluate emotional expression in English and personality traits in Arabic represents a strategic task–language alignment designed to leverage the most robust existing benchmarks in each respective language. Emotion datasets in English are abundant, well-annotated, and widely used for benchmarking expressive language, whereas Arabic resources for emotion analysis remain limited. Evaluating identical tasks across languages would introduce confounding factors related to annotation standards, label semantics, and linguistic structure; therefore, each task was evaluated within a linguistically and methodologically appropriate setting.

For both tasks, synthetic texts were generated using six large language models: GPT-4o, Gemini 1.5 Pro, Mistral Medium (v3), LLaMA 3, DeepSeek-V3, and Jais 30B. All generations were conducted between December 2024 and March 2025 via the official API endpoints of each model. A temperature value of 1.0 was applied consistently to preserve each model's inherent probability distribution and baseline stochasticity, avoiding overly constrained or deterministic outputs. All other parameters were left at their default settings to maintain standard generative behavior.

To ensure high-quality expressive data while preventing label leakage and length-based bias, prompts were designed to provide the model with a grounded target category while enforcing two key constraints. First, the models were strictly prohibited from using the category name or its direct synonyms. Second, to ensure the synthetic text mirrored the typical length of human-authored content, the models were instructed to target the median token count of the respective human baseline datasets (refer to Tables 1 and 2). This ensures that the resulting text reflects the stylistic markers of the emotion or trait rather than simple keyword repetition or length-based artifacts. For the English emotion task, the prompt template was as follows:

"Write a tweet that reflects the emotion of **[Target Emotion]**. The tweet should address general, everyday topics and be approximately **[Median Tokens]** words in length. Do not mention the name of the emotion or any direct synonyms. Convey the emotion implicitly through tone, style, and word choice."

For the Arabic personality task, the prompt was:

"اكتب نص يعكس سمة [Trait] يجب أن يتناول النص موضوعاً عاماً ومتنوعاً، وبطول تقريبي يبلغ [Median Tokens] كلمة. تنبيه هام: لا تذكر اسم السمة أو أي مرادفات مباشرة لها في النص، بل عبّر عن الشخصية من خلال أسلوب الكتابة واختيار الكلمات فقط".

By anchoring the generation to the median token count, this strategy minimizes structural bias and ensures that downstream classifiers are evaluated on stylistic and expressive signatures rather than superficial differences in text length.

Following generation, all human and AI-authored texts underwent a systematic preprocessing pipeline to isolate the stylistic signal from platform-specific noise. We removed all URLs, hashtags, and user mentions (@user) to prevent classifiers from relying on non-linguistic artifacts. Additionally, to ensure a robust stylistic footprint, we excluded all samples containing three tokens or fewer. Such short-form texts typically lack the linguistic complexity necessary for authorship attribution and can produce unreliable results in readability calculations. After this filtering, LLM-generated emotion datasets were balanced with 1,000 samples per emotion per model, while 100 samples per model were collected for each personality trait. Median token counts and full dataset summaries are reported in Tables 1 and 2.

Table 1: Summary of Human and AI-generated Datasets

| Dataset | Language | Labels | Samples (N) | Median Tokens | Source/Notes |
|---|---|---|---|---|---|
| **Ironic Corpus** | English | Irony binary | 1,950 | 18 | Wallace et al. (2014) |
| **Sarcasm Dataset** | English | Sarcasm binary | 3467 | 14 | Tanyel (2022) |
| **Text Emotion** | English | Happy, Sad | 282,782 | 21 | Cheela (2023) |
| **Emotions Dataset** | English | Happy, Sadness, Love, Rage, Fear, Surprise | 393,822 | 19 | Elgiriyewithana (2024) |
| **MSAPersonality** | Arabic | Big Five Traits | 1,335 | 46 | Chraibi et al. (2024) |

Table 2: Summary of AI-generated Datasets

| Dataset | Language | Labels | Samples (N) | Median Tokens | Source/Notes |
|---|---|---|---|---|---|
| **LLM Emotion** | English | 10 Emotions | 60,000 | **17** | 1,000 per emotion/LLM |
| **LLM Personality** | Arabic | Big Five | 3,000 | 43 | 100 per trait/LLM |

## 3.2 Experimental Pipeline

The methodology includes three major experimental pipelines: (1) AI authorship detection, (2) emotion/personality classification, and (3) linguistic feature analysis. The details of these pipelines are in Figure 1 and Figure 2.

### 3.2.1 AI Authorship Detection

In the AI authorship detection experiments, binary classifiers were trained to distinguish AI-generated from human-authored texts, using BERT for English and AraBERT for Arabic. To complement this analysis, we examined two additional conditions to probe the reliance of classifiers on surface-level textual cues. First, we evaluated a "no punctuation" condition in which all punctuation was removed from the dataset, assessing whether classifiers depend on superficial markers such as punctuation patterns. Second, we generated rewritten versions of the same texts using QuillBot (Fitria, 2021) with the standard fluency setting to evaluate the effect of paraphrasing on classifier separability. These experiments are designed to understand differences in the linguistic "fingerprints" of AI and human authors, rather than to serve as a robustness test against detection systems or to benchmark external tools. Paraphrasing experiments were conducted only for English due to the limited size of the Arabic dataset and the lack of reliable paraphrasing tools for Arabic.

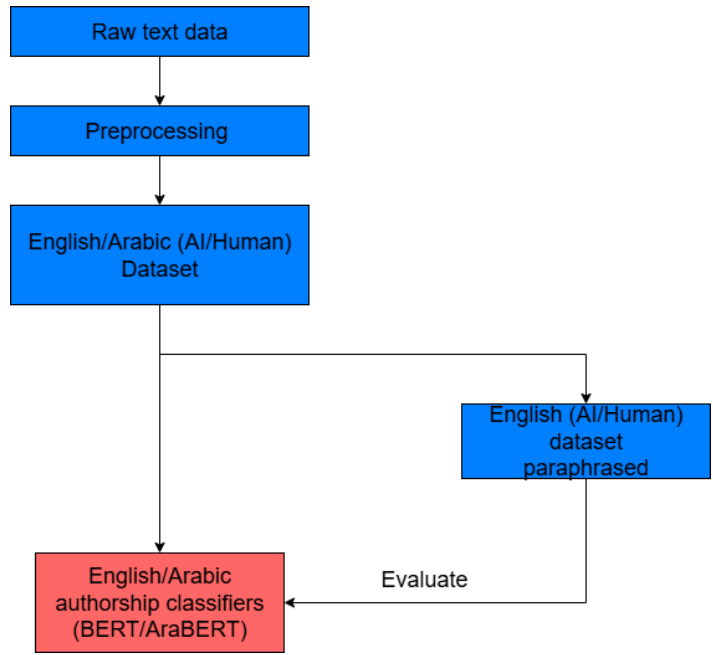

**Figure 1:** AI authorship detection

### 3.2.2 Emotion/Personality Classification

For both emotion (English) and personality (Arabic) tasks, we trained classifiers using BERT and AraBERT, respectively, which were fine-tuned for downstream emotion and personality classification tasks. We implemented three core experimental setups:

1. Training on human-authored data and testing on both human and AI samples.
2. Training on AI-generated data and testing on both human and AI samples.
3. Training on joint human–AI data and evaluating separately on each.

The rationale behind the three training setups is to test the generalizability of learned emotional/personality representations across human and AI domains. These setups aim to assess the extent to which human-authored emotional/personality cues transfer to AI-generated texts and vice versa. This helps evaluate the stylistic consistency and internal coherence of LLM outputs.

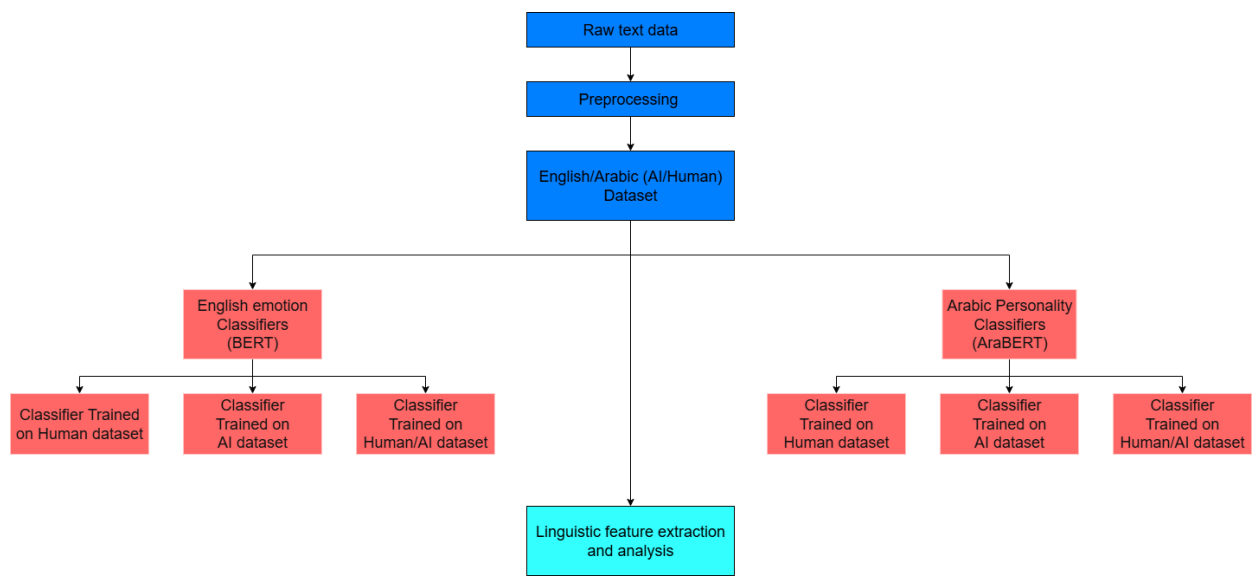

**Figure 2:** emotion/personality classification and linguistic feature analysis

### 3.2.3 Linguistic Features Analysis

To complement the classification tasks, we conducted a linguistic feature analysis in English using LIWC and several readability metrics: Flesch Reading Ease, Gunning Fog Index, Automated Readability Index, and Dale–Chall Readability Score. Descriptions of each metric are provided in Table 3. This analysis highlights stylistic and cognitive patterns that distinguish human- from AI-generated text.

For Arabic, no comparable analysis was performed. Limited sample sizes per personality trait make statistical comparisons using LIWC unreliable, and aggregating across traits could obscure meaningful trait-specific differences. Additionally, the readability metrics applied are designed for English and cannot be reliably used for Arabic. Arabic linguistic feature analysis is therefore reserved for future work, pending broader and more balanced datasets.

**Table 3:** Description of Psycholinguistic and Readability Metrics Used

| Metric | What It Measures | High Score Interpretation | Low Score Interpretation |
|---|---|---|---|
| **Tone** (Pennebaker, 2001) | Emotional valence (positive vs. negative tone) | Cheerful, positive emotional expression | Sad, anxious, angry, or negative tone |
| **Authentic** (Pennebaker, 2001) | Sincerity, honesty, and personal disclosure | Genuine, self-reflective, personal writing | Formal, distant, or impersonal tone |
| **Clout** (Pennebaker 2001) | Social status, confidence, or leadership conveyed | Confident, authoritative, high-status voice | Hesitant, insecure, deferential tone |
| **Analytic** (Pennebaker, 2001) | Logical, hierarchical, and structured thinking | Formal, academic, analytical writing | Intuitive, narrative, or conversational writing |
| **Flesch Reading Ease** (Flesch, 1948) | How easy a text is to read (0–100 scale) It's calculated using the number of words, sentences, and syllables in a text | Very easy to read (simple words, short sentences) | Very hard to read (complex sentences and vocabulary) |
| **Gunning Fog Index** (Gunning, 1952) | Years of formal education needed to understand the text, based on sentence length and the proportion of complex words | Complex writing requiring more education ( >12 = college-level) | Simple, clear writing requires less education ( <8 = general audience) |
| **Automated Readability Index (ARI)** (Senter and Smith, 1967) | Readability based on characters per word and words per sentence | Indicates higher grade level (longer words, longer sentences) | Easier to read (short words, short sentences) |
| **Dale-Chall Score** (Dale and Chall, 1948) | Readability based on use of unfamiliar or complex words | Many difficult or uncommon words; harder for general readers | Uses familiar, everyday vocabulary; easier to understand |

### 3.2.4 Model Implementation

Fine-tuning configurations for the classification models are summarized in Table 4. Parameters were selected based on grid search over learning rate, batch size, and epoch combinations.

**Table 4:** Hyperparameter Settings for Fine-Tuned BERT and AraBERT Models

| Parameter | BERT (Emotion, English) | AraBERT (Personality, Arabic) |
|---|---|---|
| Task Type | Single-label classification (10 labels) | Multi-label classification (5 binary traits) |
| Base Model | BERTForSequenceClassification | BERT-base-araBERTv2 |
| Tokenizer | BERT-base-uncased | AraBERT tokenizer |
| Max Sequence Length | 128 tokens | 128 tokens |
| Optimizer | AdamW | AdamW |
| Learning Rate | 2.00E-05 | 2.00E-05 |
| Epochs | 5 | 5 |
| Batch Size | 32 | 16 |
| Scheduler | get_linear_schedule_with_warmup | get_linear_schedule_with_warmup |
| Loss Function | Cross-entropy | Binary cross-entropy (multi-label setup) |
| Device | MPS (Apple Silicon) or CPU fallback | MPS (Apple Silicon) or CPU fallback |

## 3.3 Evaluation Metrics

All classification models were evaluated using standard performance metrics derived from confusion matrices: accuracy, precision, recall, and F1-score. To ensure statistical reliability and account for the inherent variance in social media datasets, we employed 5-fold cross-validation across all experimental conditions in Pipelines 1 and 2. This approach ensures that reported metrics reflect consistent model behavior across different data partitions rather than artifacts of a single train-test split.

Unless otherwise specified, all metrics represent the mean values across the five folds, with standard deviations indicating performance stability. Macro-averaged scores are reported to provide an aggregate measure of performance across all classes.

For the linguistic analysis in Pipeline 3, LIWC features were computed at the individual text level, and differences between human and AI-generated populations were assessed using Mann–Whitney U tests, as this non-parametric approach is robust to the non-normal distributions typical of word-category frequencies. We report findings for features that reached a significance threshold of $p < 0.05$ or those that highlight critical stylistic alignments ($p > 0.05$).

In contrast, readability metrics were calculated at the corpus level for each source. This aggregation ensures that structural ratios, such as syllables per word and words per sentence, are derived from a statistically sufficient linguistic sample, avoiding the volatility inherent in traditional readability formulas when applied to short-form texts.

# 4 Results and Discussion

This section presents and interprets the results of AI authorship detection, emotion and personality classification, LLM-Specific performance insights, and linguistic feature analysis. Evaluation focuses primarily on the F1-score, which balances precision and recall.

## 4.1 AI Authorship Detection

We trained binary classifiers to distinguish between AI-generated and human-authored texts in both English and Arabic. Results are summarized in Table 5, which reports F1-scores for the original texts as well as after removing all punctuation. Across both languages, distinguishing AI from human texts proved highly feasible, with consistently strong F1-scores. The inclusion of the no-punctuation condition provides additional insight into how the removal of punctuation affects classifier performance.

**Table 5:** AI Authorship Detection Performance (mean ± std over 5 folds)

| Language | Model | Mean F1-Score (Original) | Mean F1-Score (No Punctuation) |
|---|---|---|---|
| **English** | BERT | 0.97 ± 0.01 | 0.92 ± 0.02 |
| **Arabic** | AraBERT | 0.95 ± 0.02 | 0.93 ± 0.01 |

To examine how textual rewriting affects AI versus human separability, we paraphrased the English dataset using QuillBot and compared classifier performance before and after modification, as shown in Table 6. While both AI and human-authored texts were affected, the decline was far more pronounced for AI. Most notably, recall for AI texts dropped from 0.95 to 0.34, indicating that many AI-generated samples were no longer correctly identified after paraphrasing. In contrast, recall for human texts decreased moderately, from 0.97 to 0.88, a decline substantially smaller than that observed for AI-generated texts. The sharp drop in human precision (0.49) is a direct consequence of this recall collapse: as the classifier failed to detect paraphrased AI samples, it misclassified them as human, inflating errors in the human-labeled pool. Precision remained relatively high for AI (0.80), suggesting that when the classifier predicted a text as AI-generated, it was generally correct, but many paraphrased AI samples went undetected. This asymmetry highlights how paraphrasing can obscure stylistic cues in AI-generated text while human writing is comparatively more resilient.

Table 6: Paraphrased AI Authorship Detection Performance (mean ± std over 5 folds)

| Condition | Class | Mean Precision | Mean Recall | Mean F1-Score |
|---|---|---|---|---|
| English Original | AI | 0.95 ± 0.02 | 0.95 ± 0.01 | 0.96 ± 0.01 |
| | Human | 0.97 ± 0.01 | 0.97 ± 0.01 | |
| English Paraphrased | AI | 0.80 ± 0.02 | 0.34 ± 0.03 | 0.53 ± 0.02 |
| | Human | 0.49 ± 0.03 | 0.88 ± 0.02 | |

## 4.2 Emotion and Personality Classification

To assess the generalizability of emotion and personality classifiers across human and AI-generated texts, we trained models under three distinct regimes: using only human-authored data, only AI-generated data, and a combination of both. The same fixed test sets were used across all training conditions to ensure comparability and isolate the effect of training source. Results are summarized in Table 7.

Table 7: Classification Results for emotion/personality across human and AI training sets (mean ± std over 5 folds)

| Training Setup | Test Set | Mean F1 (Emotion) | Mean F1 (Personality) |
|---|---|---|---|
| Human | Human | **0.77 ± 0.02** | 0.45 ± 0.04 |
| | AI | 0.18 ± 0.04 | 0.54 ± 0.03 |
| AI | Human | 0.21 ± 0.03 | 0.51 ± 0.03 |
| | AI | **0.77 ± 0.02** | 0.83 ± 0.01 |
| Human + AI | Human | 0.71 ± 0.02 | **0.66 ± 0.03** |
| | AI | 0.73 ± 0.02 | **0.87 ± 0.02** |

In the English emotion classification pipeline, models trained on human data performed well when tested on human-authored texts but exhibited sharp performance declines on AI-generated samples. Conversely, training on AI data produced strong performance on AI test samples but weak performance on human texts. Combining human and AI-generated training data did not lead to significant improvements; rather, model performance declined across both subsets. This suggests that the emotional cues conveyed by LLMs differ stylistically and semantically from those expressed by humans, thereby limiting cross-source generalization in a high-resource language. Consequently, incorporating AI-generated data into the training set appears to introduce noise rather than provide beneficial complementary information.

In contrast, the Arabic personality classification task exhibited a different pattern. Training solely on human-authored data resulted in poor performance on both test sets. When trained exclusively on AI-generated data, the model's performance improved substantially on the AI test set and showed moderate gains on human-authored samples. The most pronounced improvement in generalization was observed when training on the combined human and AI dataset, yielding enhanced performance across both test sets, with the largest gains occurring on human-authored texts.

Taken together, these results indicate that augmenting under-resourced language datasets with LLM-generated samples can substantially improve classification performance, especially on human-authored Arabic texts. This contrasts with the English emotion classification findings, where training on synthetic data alone did not enhance generalization.

Notably, the performance gains observed in Arabic likely reflect broader challenges inherent to under-resourced languages rather than characteristics specific to personality traits. This interpretation aligns with findings from prior Arabic emotion classification studies (Alyammi, 2025), where synthetic data augmentation similarly resulted in marked improvements.

## 4.3 LLM-Specific Performance Insights

To better understand the performance of different LLMs across classification tasks, we conducted a comparative analysis of LLM outputs. Given the large number of labels and models involved, reporting all per-class metrics would be excessive; therefore, we adopt a selective reporting strategy that emphasizes key trends. Specifically, we highlight average macro F1-scores per LLM as a general performance indicator, alongside the identification of top- and bottom-performing models within each category, depending on the classification task. This approach aims to reveal model-emotion or model-trait combinations where generative alignment with human-authored text is particularly strong or notably limited. The following subsections present these findings separately for the emotion and personality classification tasks.

### 4.3.1 LLM-Specific Trends in Emotion Expression

In addition to the overall trends observed across experimental conditions, we identified notable variation in how different LLMs approximate human-like emotional expression. Although average performance was comparable across models, as reported in Table 8, certain LLMs demonstrated stronger alignment with specific emotions. Table 9 presents the top-performing LLMs per emotion, highlighting instances where AI-generated texts conveyed emotional tones more consistent with human-authored texts. Importantly, these scores do not necessarily reflect strong absolute performance; rather, they indicate relative strengths in contexts where some models substantially outperformed others.

**Table 8:** Average macro F1-score per LLM in Human-to-AI classification (mean ± std over 5 folds)

| LLM | GPT-4o | Gemini | DeepSeek | Mistral | LLaMA | Jais |
|---|---|---|---|---|---|---|
| **Mean Macro F1 Score** | 0.14 ± 0.03 | **0.18 ± 0.03** | 0.17 ± 0.04 | 0.13 ± 0.03 | 0.12 ± 0.04 | 0.12 ± 0.05 |

Despite similar averages, model-specific peaks are noteworthy. For instance, GPT-4o and Gemini outperformed others in disgust, while DeepSeek and Gemini showed strength in irony. Mistral demonstrated higher F1-scores in anticipation and trust, and LLaMA was the highest for Sadness. These findings indicate that some LLMs are better aligned with specific emotional styles, which have implications for emotion modeling and alignment strategies.

**T Table 9:** Top-performing LLMs per emotion in Human-to-AI classification (F1 mean ± std over 5 folds)

| Emotion | Top LLM(s) | F1 Score(s) |
|---|---|---|
| Disgust | GPT-4o, Gemini | 0.75 ± 0.02, 0.66 ± 0.03 |
| Irony | DeepSeek, Gemini | 0.60 ± 0.02, 0.54 ± 0.02 |
| Anticipation | Mistral | 0.40 ± 0.01 |
| Trust | Mistral | 0.41 ± 0.01 |
| Surprise | — | — |
| Joy | — | — |
| Fear | — | — |
| Sadness | LLaMA | 0.49 ± 0.01 |
| Anger | — | — |
| Sarcasm | DeepSeek, Jais | 0.39 ± 0.01, 0.32 ± 0.01 |

However, the uniformly low F1-scores observed for several emotions highlight the limited generalizability of emotional cues from human-authored data to AI-generated text. This limitation does not imply model failure or flawed data. In fact, when classifiers are trained and tested within the same domain, performance improves substantially, indicating that while LLMs exhibit consistent internal cues, these cues differ stylistically from those found in human writing.

In stark contrast to the results obtained from training on human-authored texts, classifiers trained and tested solely on AI-generated data achieved substantially higher F1-scores across all models and emotions. This improvement indicates that LLMs encode emotion-specific stylistic cues that are both consistent and distinguishable within their own outputs, even though these cues diverge from the ways humans express emotion in text.

Table 10 presents the average macro F1-score for each LLM across all emotion categories under the within-domain (AI-to-AI) condition. The results indicate consistently strong performance across models, with Gemini and GPT-4o emerging as the most reliable generators of emotionally distinguishable text. Notably, LLaMA, followed by Mistral, exhibited the lowest average F1-scores among the models, suggesting potential limitations in their ability to encode affective variation compared to other LLMs.

**Table 10:** Average macro F1-score per LLM in AI-to-AI classification (mean ± std over 5 folds)

| LLM | GPT-4o | Gemini | DeepSeek | Mistral | LLaMA | Jais |
|---|---|---|---|---|---|---|
| **Macro F1** | **0.95 ± 0.01** | 0.93 ± 0.01 | 0.87 ± 0.02 | 0.83 ± 0.02 | 0.72 ± 0.02 | 0.87 ± 0.02 |

To further examine inter-model variation, Table 11 highlights the lowest-performing LLM for each individual emotion. While these scores remain substantially higher than those observed in the human-to-AI setting, certain model–emotion combinations consistently underperformed relative to their peers. In particular, LLaMA exhibited the weakest results across most emotions, followed by Mistral, whereas GPT-4o and Gemini consistently ranked among the top performers. These discrepancies may reflect differences in model alignment, sensitivity to subtle affective cues, or lexical diversity within generated outputs.

**Table 11:** Worst-performing LLMs per emotion in AI-to-AI classification (F1 mean ± std over 5 folds)

| Emotion | Lowest LLM(s) | F1 Score(s) (mean ± std) |
|---|---|---|
| **Disgust** | LLaMA, Jais | 0.80 ± 0.02, 0.81 ± 0.01 |
| **Irony** | LLaMA, Mistral | 0.57 ± 0.01, 0.75 ± 0.02 |
| **Anticipation** | LLaMA | 0.81 ± 0.02 |
| **Trust** | LLaMA | 0.65 ± 0.01 |
| **Surprise** | LLaMA, Mistral | 0.71 ± 0.02, 0.73 ± 0.02 |
| **Joy** | Mistral, LLaMA | 0.64 ± 0.01, 0.68 ± 0.02 |
| **Fear** | Mistral, DeepSeek | 0.79 ± 0.02, 0.82 ± 0.01 |
| **Sadness** | DeepSeek, LLaMA | 0.78 ± 0.02, 0.79 ± 0.03 |
| **Anger** | LLaMA, DeepSeek | 0.66 ± 0.02, 0.66 ± 0.01 |
| **Sarcasm** | Mistral, LLaMA | 0.58 ± 0.03, 0.62 ± 0.02 |

### 4.3.2 LLM-Specific Trends in Personality Expression

Compared to the emotion classification task in English, the personality classification task in Arabic presents additional challenges stemming from the limited size and variability of the dataset. As a result, we refrain from reporting trait-specific performance across LLMs, since small sample sizes per class make fine-grained comparisons statistically unreliable. Instead, we focus on average macro F1-scores per model, which provide a more robust indication of each LLM's ability to emulate human-like personality traits in aggregate.

Table 12 reports the average F1-scores per LLM when trained on human-authored texts and tested on AI-generated data. Although overall performance was low across the board, Mistral emerged as the strongest performer, achieving the highest average score, followed by Jais, a model designed specifically for Arabic, while LLaMA consistently underperformed. The strong performance of Mistral is notable, especially considering that it was not explicitly optimized for Arabic content. Jais's competitive performance is consistent with its regional alignment, further underscoring the importance of language specialization.

**Table 12:** Average macro F1-score per LLM in Human-to-AI classification (mean ± std over 5 folds)

| LLM | GPT-4o | Gemini | DeepSeek | Mistral | LLaMA | Jais |
|---|---|---|---|---|---|---|
| **Macro F1 Score** | 0.50 ± 0.03 | 0.42 ± 0.04 | 0.39 ± 0.03 | **0.53 ± 0.02** | 0.29 ± 0.05 | 0.50 ± 0.03 |

Table 13 presents the results under the AI-to-AI training and testing condition. Overall performance improves substantially across all LLMs, with Gemini and Mistral leading, followed closely by Jais and GPT-4o. LLaMA remained the lowest-performing model, consistent with its results in the human-to-AI condition. These findings reinforce the conclusion that certain LLMs, particularly Mistral, demonstrate stronger stylistic consistency and internal coherence in Arabic personality generation, even without fine-tuning.

**Table 13:** Average macro F1-score per LLM in AI-to-AI classification (mean ± std over 5 folds)

| LLM | GPT-4o | Gemini | DeepSeek | Mistral | LLaMA | Jais |
|---|---|---|---|---|---|---|
| **Macro F1 Score** | 0.82 ± 0.02 | **0.86 ± 0.02** | 0.80 ± 0.03 | 0.83 ± 0.02 | 0.75 ± 0.03 | 0.81 ± 0.02 |

These findings suggest that while synthetic personality data is more internally consistent than its human-authored counterpart, variability across LLMs remains. This performance gradient is noteworthy given the models' differing levels of regional optimization. Jais, for instance, is explicitly tuned for Arabic, while Mistral is not. Mistral's strength across both training conditions may reflect more robust generalization capabilities, while Jais's competitive performance highlights the value of language-specific alignment.

However, the limited dataset size restricts the statistical confidence of the model-level comparisons and precludes strong conclusions about performance differences among individual models. Future studies with larger, more balanced Arabic personality datasets are needed to validate these findings and to determine whether observed patterns stem from language-specific factors or from differences in how LLMs encode personality.

## 4.4 Linguistic Feature and LIWC Analysis

To further probe stylistic and psycholinguistic differences between human- and AI-authored texts, we conducted a linguistic feature analysis using both LIWC-derived metrics and five readability metrics (as described in Section 3). These interpretable features complement black-box classifier results by highlighting surface-level and cognitive disparities in expression. This analysis was conducted across English emotion categories to identify the underlying linguistic 'fingerprints' of the models. Only features with statistically significant differences are reported: $p < 0.05$ indicates a meaningful divergence from human texts, while $p > 0.05$ highlights features closely aligned with human writing.

For LIWC metrics (Table 14), we focus on features that are either clearly divergent from human texts (far from human baseline, $p < 0.05$) or closely aligned with human writing (near human baseline, $p > 0.05$). Given the large volume of results across all emotions and models, this selection provides a concise overview of the most informative stylistic distinctions.

**Table 14:** Emotion-Level Linguistic Divergence in English

| Emotion | Source | Analytic | Clout | Authentic | Tone |
|---|---|---|---|---|---|
| **Anger** | Human | 20.19 | 2.51 | 98.18 | 1.00 |
| | GPT-4o | 4.73 | 41.67 | 92.11 | 93.14 |
| | DeepSeek | 1 | 1 | 89.77 | 94.25 |
| **Disgust** | Human | 58.10 | 26.56 | 41.91 | 16.76 |
| | Gemini | 83.86 | 34.59 | 7.73 | 1.16 |
| | Mistral | 6.74 | 1.00 | 61.22 | 1.00 |
| **Irony** | Human | 66.39 | 51.86 | 31.99 | 23.97 |
|  | GPT-4o | 1.16 | 3.03 | 90.17 | 99.00 |
|  | Mistral | 92.08 | 83.40 | 88.17 | 93.67 |
| **Sadness** | Human | 18.71 | 1.60 | 99.00 | 1.00 |
|  | Jais | 34.82 | 1.00 | 99.00 | 3.45 |
|  | DeepSeek | 47.48 | 1 | 99 | 96.42 |
| **Joy** | Human | 25.12 | 3.34 | 98.71 | 91.90 |
|  | GPT-4o | 94.66 | 18.64 | 88.94 | 96.76 |
|  | Mistral | 99.00 | 49.70 | 52.89 | 99.00 |
| **Anticipation** | Human | 77.77 | 48.86 | 23.03 | 48.85 |
|  | GPT-4o | 79.43 | 36.15 | 98.91 | 24.65 |
|  | Gemini | 6.27 | 1.00 | 99.00 | 99.00 |
| **Fear** | Human | 22.65 | 2.87 | 99.00 | 1.00 |
|  | GPT-4o | 47.13 | 1.00 | 99.00 | 30.58 |
|  | Gemini | 6.64 | 1.00 | 97.32 | 39.14 |
| **Irony** | Human | 66.39 | 51.86 | 31.99 | 23.97 |
|  | GPT-4o | 1.16 | 3.03 | 90.17 | 99.00 |
|  | Gemini | 68.74 | 7.03 | 92.47 | 36.60 |
| **Sarcasm** | Human | 44.52 | 42.64 | 37.29 | 65.94 |
|  | DeepSeek | 14.75 | 1 | 90.8 | 99 |
|  | Gemini | 67.34 | 3.28 | 97.86 | 98.30 |
| **Surprise** | Human | 23.04 | 3.37 | 99 | 59.22 |
|  | Jais | 96.07 | 11.11 | 77.35 | 94.1 |
|  | GPT-4o | 1 | 1 | 99 | 49.51 |

Several consistent trends emerged. Across nearly all emotions, AI-generated texts exhibited inflated Authenticity scores, often exceeding 90, suggesting a tendency toward overly self-disclosing or "sincere" sounding language. In contrast, human-authored texts demonstrated greater stylistic variability and more moderate Authenticity, consistent with natural expressive variation.

Marked differences emerged in Tone, a proxy for emotional valence, particularly in the outputs of DeepSeek and GPT-4o. In negative emotion categories such as anger, fear, and sadness, human texts consistently scored near zero, reflecting an appropriately negative emotional tone. In contrast, GPT-4o and DeepSeek produced unexpectedly high Tone values, signaling a misalignment in affective realism. Sarcasm and surprise were also poorly approximated: AI-generated sarcasm frequently carried excessively high Tone and low Clout, lacking the subtlety seen in human writing, while surprise was expressed either with inflated formality (e.g., Jais) or flat delivery (e.g., GPT-4o).

Analytic scores, which reflect logical structure and syntactic complexity, further differentiated models. While human text showed mid-range values across emotions, LLMs diverged: Gemini, Jais, and Mistral often produced highly structured outputs, whereas GPT-4o and DeepSeek leaned more conversational or fragmented. These findings suggest that LLMs exhibit internally consistent yet stylized patterns that diverge from the nuanced and affectively grounded styles found in human-authored text. The cumulative effect is that while some models may appear affectively coherent, they often do so through over-simplified or exaggerated stylistic cues rather than authentically mirroring human expression.

**Table 15:** Corpus-Level Readability Metrics by Source

| Source | Flesch Reading Ease | Gunning Fog | ARI | Dale-Chall |
|---|---|---|---|---|
| Human | 72.1 | 9.8 | 9.5 | 8.5 |
| GPT-4o | 87.6 | 3.9 | 4.7 | 1.0 |
| DeepSeek | 92.2 | 3.6 | 3.7 | 0.6 |
| Gemini | 81.3 | 4.8 | 5.3 | 2.5 |
| Jais | 78.6 | 5.1 | 5.9 | 5.3 |
| LLaMA | 82.5 | 5.3 | 6.4 | 3.2 |
| Mistral | 85.5 | 3.9 | 5.0 | 1.0 |

Table 15 reports corpus-level readability metrics computed separately for human-authored text and for each LLM. Aggregating readability measures at the corpus level allows structural ratios (e.g., sentence length and word complexity) to be estimated from sufficiently large and continuous text, for which traditional readability formulas are designed.

Overall, human-written text exhibits consistently higher linguistic complexity than AI-generated text. Human writing achieves a Flesch Reading Ease score of 72.1, whereas most LLMs (e.g., DeepSeek: 92.2; GPT-4o: 87.6) produce substantially more readable outputs, indicating simpler vocabulary and shorter sentence structures. This pattern is reinforced by the Dale–Chall Readability Score, where human text averages 8.5, while several models (e.g., DeepSeek, GPT-4o, and Mistral) score at or below 1.0, reflecting reliance on very common lexical choices.

Grade-level metrics further highlight this disparity. Human text averages 9.8 on the Gunning Fog Index, corresponding to secondary-school reading levels, whereas most LLMs generate text aligned with elementary to early middle-school complexity (e.g., DeepSeek: 3.6; GPT-4o and

Mistral: 3.9). Although Jais (5.1) and LLaMA (5.3) approach human complexity more closely, they remain less lexically dense and structurally varied.

A similar trend is observed for the Automated Readability Index (ARI). Human writing attains an ARI of 9.5, consistent with late high-school or early college-level text, while all LLMs yield markedly lower values (ranging from 3.7 to 6.4). Collectively, these results indicate that LLM-generated text prioritizes accessibility and fluency at the expense of syntactic and lexical complexity, underscoring a persistent stylistic distinction between human and synthetic writing.

# 5 Conclusion

This study investigated the extent to which LLMs can emulate human-like emotional expression and personality traits in generated text, focusing on English for emotion and Arabic for personality. Through a combination of neural classification and linguistic feature analysis, we assessed whether human-labeled classifiers could accurately detect emotion and personality traits in AI-generated content and whether AI text could be reliably distinguished from human-authored text.

Across both languages, binary classifiers consistently distinguished between human and AI-generated texts with high F1-scores, confirming that current LLMs still exhibit detectable stylistic differences from human writing. While earlier experiments showed that removing punctuation moderately affected classification, the decline observed under paraphrasing, particularly for AI-authored texts, was far more pronounced, indicating that classifiers rely on broader superficial cues beyond simple formatting. Performance sharply declined when inputs were paraphrased, particularly for AI-authored texts. This suggests that classifiers rely on superficial linguistic cues that can be easily obfuscated, raising concerns about the long-term viability of detection strategies in the face of increasingly fluent and malleable generation tools.

In the emotion classification task, results revealed that classifiers trained on human-authored data failed to generalize effectively to AI-generated samples and vice versa. While human-labeled training data produced strong results on human test sets, performance dropped significantly when tested on AI outputs. Similarly, models trained on AI-authored data failed to capture human emotional tone. Interestingly, combining human and AI data in training did not consistently improve results in English, whereas in Arabic it provided modest gains for personality classification, potentially reflecting both language-specific modeling challenges and resource differences. This discrepancy may reflect language-specific limitations and resource imbalances.

Further, LLM-wise comparisons revealed substantial variation in how well models captured affective signals. In the English emotion task, models like GPT-4o and Gemini produced more human-aligned outputs for certain emotions (e.g., disgust), while others (e.g., LLaMA) performed poorly across most categories. In Arabic, Mistral and Jais demonstrated relatively stronger alignment with personality cues, despite the limited size of training data.

Linguistic feature analysis revealed clear stylistic differences between human and LLM-generated texts in the English emotion dataset. Using LIWC (at the individual sample level) and corpus-level readability metrics, human-authored samples exhibited greater linguistic complexity, reflected in higher Gunning Fog and Dale-Chall scores, and distinct psycholinguistic profiles, particularly in authenticity and emotional tone. LLM outputs, while often more readable, tended to favor analytic over narrative styles and varied in how closely they mirrored human expression across different emotions. These findings highlight the value of feature-based analysis as a complementary lens, assessing not only classification performance but also the degree to which models emulate human-like linguistic patterns.

Taken together, these findings highlight that while LLMs are capable of generating emotionally or psychologically structured text, these outputs often fail to mirror the nuanced linguistic signatures found in human-authored writing. The inconsistency across models and languages underscores the need for more refined generation, evaluation, and alignment techniques, especially for under-resourced languages and culturally embedded constructs like personality.

## 5.1 Limitations and Future Work

While this study offers a comprehensive analysis, certain boundaries naturally emerge. The Arabic personality dataset, for instance, was limited in size, constraining both trait-level and stylistic generalization. Similarly, while our linguistic analyses considered features such as LIWC and readability, certain structural aspects, such as punctuation or text preprocessing choices, may influence classifier sensitivity, which warrants further exploration. Moreover, while emotion and personality were studied independently, their real-world interplay warrants future investigation. Some observed divergences may also reflect broader linguistic or cultural patterns rather than model deficiencies. Additionally, while our prompting strategy aimed to minimize topical bias by encouraging models to select diverse subjects, the potential for latent topical alignment between specific traits and certain domains remains a challenge in synthetic data generation.

Future work could expand on this direction, particularly by developing or adapting tools suited for non-English text, as our analysis in Arabic was constrained by both dataset size and the lack of comparable linguistic metrics. Subsequent studies should explore more rigorous topical-control frameworks to ensure that classifiers distinguish between stylistic markers and subject-specific signals with even greater precision.

Furthermore, research could explore the impact of augmenting training data with AI-generated text across different languages and resource settings. While we observed improvements in Arabic (an under-resourced language), the same approach did not benefit English, suggesting that the utility of synthetic augmentation is highly context dependent. Broader cross-linguistic studies are needed to identify when and where such augmentation strategies are effective.

Exploration of advanced alignment techniques at generation time, such as stylistic steering or controllable decoding, may help bring synthetic outputs closer to human norms. Likewise, adversarial fine-tuning, where a discriminator pressures the model to produce less detectable text, could reduce stylistic gaps. Finally, detection frameworks should increasingly rely on deeper semantic and discourse-level features, beyond surface-level stylistics, to ensure robustness and transparency in generative AI deployments.

## Statements and Declarations

**Acknowledgement:** The authors would like to thank King Saud University, Riyadh, Saudi Arabia for supporting the work by Ongoing Research Funding program (ORF-2026-846), King Saud University, Riyadh, Saudi Arabia

**Competing interest:** The authors have no competing interests to declare that are relevant to the content of this article.

**Data**: the dataset used is available upon request.